\documentclass[10pt,twocolumn,letterpaper]{article}

\usepackage{wacv}
\usepackage{times}
\usepackage{epsfig}
\usepackage{graphicx}
\usepackage{amsmath}
\usepackage{amssymb}

\usepackage{booktabs}
\usepackage{pifont}
\usepackage{multirow}
\usepackage{enumitem}

\def\wacvPaperID{700} %

\wacvfinalcopy %

\ifwacvfinal
\def\assignedStartPage{1} %
\fi

\newcommand{\PAR}[1]{\vskip1pt \noindent {\bf #1~}}
\newcommand{\PARbegin}[1]{\noindent {\bf #1~}}

\ifwacvfinal
\usepackage[breaklinks=true,bookmarks=false]{hyperref}
\else
\usepackage[pagebackref=true,breaklinks=true,colorlinks,bookmarks=false]{hyperref}
\fi

\ifwacvfinal
\else
\fi

\makeatletter
\def\@fnsymbol#1{\ensuremath{\ifcase#1\or \dagger\or \ddagger\or
   \mathsection\or \mathparagraph\or \|\or **\or \dagger\dagger
   \or \ddagger\ddagger \else\@ctrerr\fi}}
\makeatother

\begin{document}

\title{Reducing the Annotation Effort for Video Object Segmentation Datasets}

\author{
  \hspace{-0.4cm}
  \begin{tabular}[t]{c}
    Paul Voigtlaender$^1$ \quad Lishu Luo$^{2,}$\thanks{Work performed at RWTH Aachen under a RWTH Aachen - Tsinghua Junior Research Fellowship.} \quad Chun Yuan$^2$ \quad Yong Jiang$^2$ \quad Bastian Leibe$^1$ \\
    $^1$RWTH Aachen University $^2$Tsinghua University\\
    {\tt\small \{voigtlaender,leibe\}@vision.rwth-aachen.de \quad \tt\small \{lls18@mails,yuanc@sz,jiangy@sz\}.tsinghua.edu.cn}
\end{tabular}
}

\maketitle

\begin{abstract}
For further progress in video object segmentation (VOS), larger, more diverse, and more challenging datasets will be necessary. However, densely labeling every frame with pixel masks does not scale to large datasets. We use a deep convolutional network to automatically create pseudo-labels on a pixel level from much cheaper bounding box annotations and investigate how far such pseudo-labels can carry us for training state-of-the-art VOS approaches. A very encouraging result of our study is that adding a manually annotated mask in only a single video frame for each object is sufficient to generate pseudo-labels which can be used to train a VOS method to reach almost the same performance level as when training with fully segmented videos.
We use this workflow to create pixel pseudo-labels for the training set of the challenging tracking dataset TAO, and we manually annotate a subset of the validation set. Together, we obtain the new TAO-VOS benchmark, which we make publicly available at \url{www.vision.rwth-aachen.de/page/taovos}. While the performance of state-of-the-art methods on existing datasets starts to saturate, TAO-VOS remains very challenging for current algorithms and reveals their shortcomings.
\end{abstract}

\vspace{-4mm}
\section{Introduction}
Semi-supervised Video Object Segmentation (VOS) is the task of segmenting objects in every frame of a video given their ground truth masks in the first frame in which they appear. VOS is a fundamental task in computer vision with important applications including video editing, robotics, and self-driving cars. Recently, VOS has received a lot of attention in the computer vision community and the quality of results has significantly improved. In addition to the adoption of recent deep learning methods, this progress was largely driven by the introduction of the DAVIS 2017 \cite{DAVIS2017} and the YouTube-VOS \cite{Xu18Arxiv} benchmarks. 

With its size of 60 videos for training and 30 videos for validation, DAVIS 2017 is very useful for measuring the performance of VOS methods, but its utility for training is limited. 
The introduction of the much larger YouTube-VOS dataset with 3,471 training sequences was a milestone for VOS and led to a large jump in result quality. In particular, it enabled the development of end-to-end trained VOS methods. However, the videos of YouTube-VOS are short, contain relatively low diversity with a focus mainly on persons and animals, the number of objects per video is relatively low, and the total amount of data is still much smaller than what is available for other areas of computer vision such as image classification or object detection. 

Hence, we argue that in order to make significant progress in VOS, new datasets which contain more diverse data, longer sequences, and more objects per sequence will be necessary. However, manually labeling all objects in a video with segmentation masks for every frame is extremely time consuming, which limits the creation of such datasets.

\PAR{Contributions.} Towards the goal of creating VOS datasets with less effort, we make the following contributions.

\textbf{(1)} We show that densely labeling all frames in a video with segmentation masks by hand is not necessary to benefit from the additional data. Only annotating bounding boxes requires an order of magnitude less effort, and yet we show that these labels can be used to train a VOS model and achieve results which come close to what can be achieved with dense pixel-wise labeling.
\textbf{(2)} We investigate a strategy of manually annotating a single segmentation mask per object (in a single frame), resulting in a huge reduction of the number of masks to be annotated. We evaluate this methodology on the YouTube-VOS dataset. Here, we only use 4\% of all hand-labeled masks and are able to retain more than 97\% of the performance that is achieved with full supervision, which demonstrates that this setup achieves an excellent trade-off between annotation effort and utility of the labels for training.
\textbf{(3)} As a first step towards new VOS datasets, we apply this method for labeling the recent Tracking Any Object (TAO) dataset \cite{Dave20Arxiv}, which so far only had bounding box annotations, and obtain the challenging TAO-VOS benchmark, which we will make publicly available.
\textbf{(4)} We evaluate current state-of-the-art methods on the new TAO-VOS benchmark and show that TAO-VOS is effective at highlighting significant performance differences, something that is no longer possible for the already saturated DAVIS and YouTube-VOS benchmarks.

\section{Related Work}
In the following, we review work on learning with fewer segmentation labels and reducing annotation effort.

\PAR{Weakly-Supervised Image Segmentation.} In the context of semantic segmentation or instance segmentation for static images, most focus has been put on learning segmentation only from image-level classification labels \cite{Kolesnikov16ECCV, Ahn18CVPR, Ahn19CVPR}.

Other methods closer to our work focus on segmentation from bounding box supervision \cite{Dai15ICCV, Khoreva17CVPR, Hsu19NIPS}. However, even when using bounding boxes, the result quality still lacks behind what can be achieved with more supervision.

In contrast to the aforementioned methods, we consider videos with given bounding box annotations and we also make use of existing segmentation annotations for static image datasets, allowing us to create segmentation masks of much higher quality.

\PAR{Weakly-Supervised VOS.} Recent work on weakly-supervised VOS mainly focused on learning VOS only from unlabeled videos \cite{Lai19BMVC, Li19NIPS, Lu20CVPR, Lai20CVPR, Wang19ICCV, Yang19ICCV_AD, Lu20CVPR}. %
 While being scientifically interesting, these approaches still leave a big gap to the results which can be achieved with full supervision. In our work, we make use of more supervision and get very close to the results of fully-supervised methods.

\PAR{Reducing the Effort of Pixel-wise Annotations.}
There are many methods for semi-automatic image-level segmentation, which can be used to speed up the annotation process. For example, user input in the form of scribbles \cite{Rother04Siggraph}, bounding boxes \cite{Xu17BMVC}, or clicks \cite{Xu16CVPR, Mahadevan18BMVC, Jain19IJCV} can be used to interactively segment objects. Other examples are automatically predicted polygons \cite{Castrejon17CVPR, Acuna18CVPR} or segments \cite{Andriluka18Arxiv}, which can be manipulated by the user.
While speeding up the annotation for static images, most of these methods do not make use of existing bounding box level annotations and we found them often not to be robust enough to produce good results on out of domain data.

Recently, Voigtlaender \etal~\cite{Voigtlaender19CVPR_MOTS} introduced a semi-automatic annotation procedure to create dense segmentation labels for videos based on bounding box tracking annotations. Similar to our work, they use a convolutional network which converts bounding boxes into segmentation masks. However, they considered much smaller datasets and invested significant manual interaction in order to remove any errors in the annotations. Instead, we focus on a much larger dataset and allow some remaining errors in the annotations. This allows us to significantly reduce the amount of manual effort required, and we show that the remaining errors only have a minor impact on the final performance of models trained using these labels.

\begin{figure*}[t]
\begin{center}
\includegraphics[width=0.23\textwidth]{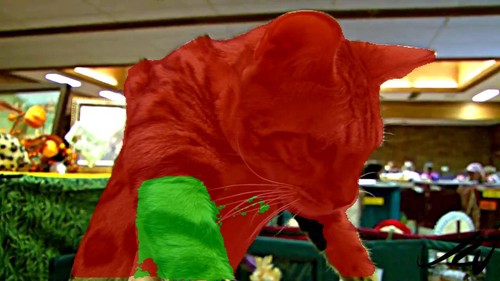}
\includegraphics[width=0.23\textwidth]{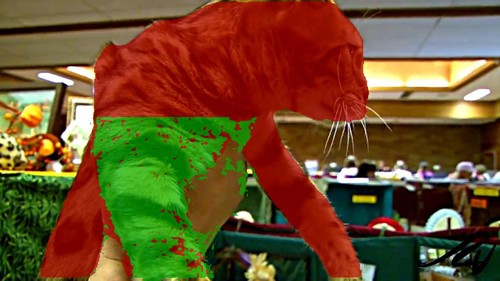}
\includegraphics[width=0.23\textwidth]{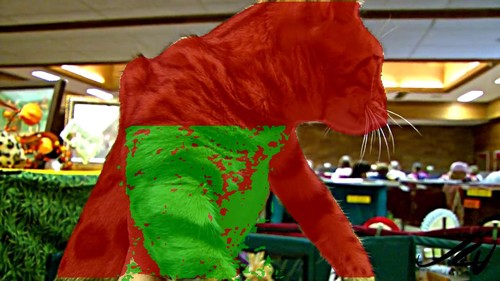}
\includegraphics[width=0.23\textwidth]{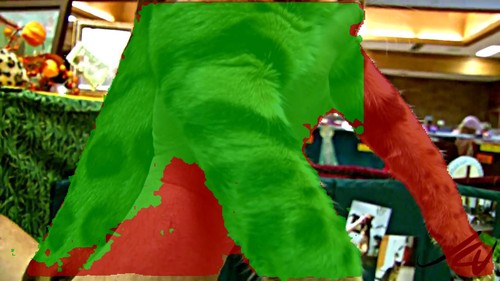}
\\
\vspace{1mm}
\includegraphics[width=0.23\textwidth]{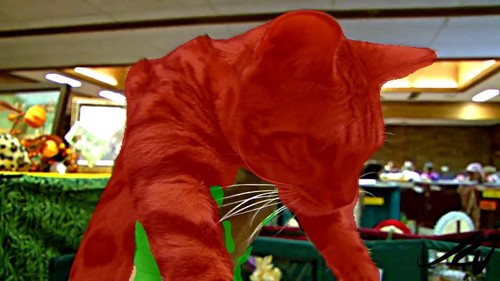}
\includegraphics[width=0.23\textwidth]{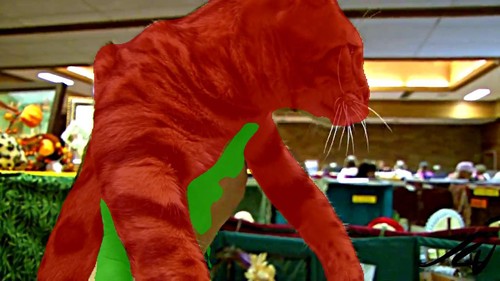}
\includegraphics[width=0.23\textwidth]{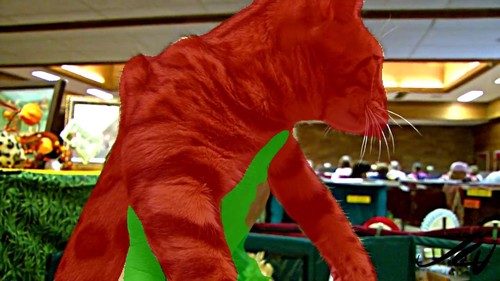}
\includegraphics[width=0.23\textwidth]{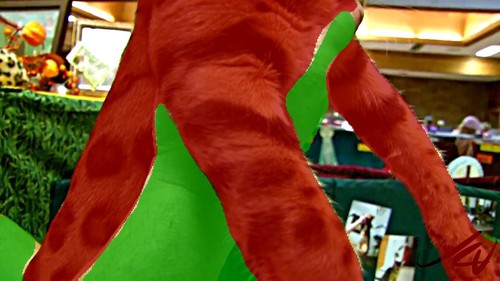}
\end{center}
\caption{\label{fig:pseudo-labels-qualitative}Qualitative example of automatically obtained pseudo labels on the YouTube-VOS training set. The first row shows the results produced by Box2Seg without using any manual mask annotations. The second row shows the result after fine-tuning on a single annotated frame per object. It can be seen that the segmentation quality improves significantly by only adding one mask per object manually. Note that this is a hard example and the mask quality looks better for most sequences.}
\end{figure*}

\section{Reducing the Annotation Effort for VOS}
Annotating dense segmentation labels for videos is extremely expensive. However, just annotating bounding boxes is much less effort and has already been done for more datasets like TAO \cite{Dave20Arxiv}. In this paper, we evaluate which level of results can be achieved with a reduced set of annotations by automatically converting bounding boxes into segmentation masks. Furthermore, we investigate how the usefulness for training of automatically created pseudo-labels can be further improved.

\subsection{Creating Pseudo-Labels from Boxes}
\label{sec:creating-pseudo-labels}

When annotating segmentation labels for static images, there is much less redundancy, since there is a much larger variety between different images than between different frames of the same video, and segmentation labels for static image datasets like COCO \cite{coco} are readily available.
We hypothesize that the information from already labeled static image datasets can be automatically transferred to videos for which only bounding boxes are labeled. In order to investigate how well this works, we adopt the powerful Box2Seg method for converting bounding boxes into segmentation masks  \cite{Luiten18ACCV}.

\PAR{Box2Seg.} The Box2Seg network has also been used successfully in other works \cite{Luiten18ACCV, Voigtlaender19CVPR_MOTS, Voigtlaender20CVPR}. Box2Seg is a fully-convolutional network based on the DeepLabv3+ \cite{Chen18ECCV} segmentation architecture which uses an Xception-65 \cite{CholletCVPR17} backbone. 
The input to Box2Seg is an image together with a bounding box. The image is concatenated with a fourth input channel which encodes the bounding box as a mask. Afterwards, all four input channels are cropped to the box with an added margin, and then resized to a fixed size of $385\times385$ pixels. We train Box2Seg on COCO to predict the pixel-precise mask of an object given its bounding box. %

\PAR{Fine-Tuning.}
Just using a bounding box as input, Box2Seg already produces pseudo-labels of high quality as will be shown in Sec.~\ref{sec:experiments}. However, for some applications it might be necessary to get even better labels without resorting to extensive labeling of every frame. 
We can annotate a single mask (in a single frame) for each object and then use this mask to fine-tune Box2Seg on it for 300 steps before creating the masks for this object on the other frames. We hypothesize that this should achieve a good effort/quality trade-off.
We evaluate this methodology on the YouTube-VOS 2018 \cite{Xu18Arxiv} dataset. Here, on average 25 masks are annotated for each object. Hence, by annotating only a single mask, the effort is reduced 25-fold, and yet using the resulting labels to train, a result which is very close to using all annotations can be achieved (\cf Sec.~\ref{sec:experiments}).

In the usual semi-supervised VOS task, always the first frame in which an object appears is given as annotated ground truth. However, using the first frame is a sub-optimal choice, and using instead the middle frame is a better heuristic \cite{Griffin20CVPR}. Hence, for each object we take the middle frame from all frames in which this object appears. The improvement of the quality of pseudo-labels through fine-tuning is illustrated in Fig.~\ref{fig:pseudo-labels-qualitative}.

\subsection{Training using Pseudo-Labels}
\label{sec:train-with-pseudo-labels}

After obtaining the pseudo-labels from Box2Seg, we use them to train a VOS method on them. Here, we want to investigate what performance can be achieved by performing the training using the obtained pseudo-labels instead of the manually annotated ground truth.

As a case study, we use the extremely successful space-time memory network (STM-VOS) \cite{Oh19ICCV} architecture. STM-VOS uses two networks. 
The first network is a memory encoder, which takes an image together with a segmentation mask as input and produces key and value embeddings used as memory to segment further frames. This mask used by the memory encoder can be either given as first-frame ground truth, or it can be the prediction of STM-VOS for a previous frame.
The second network consists of a query encoder, a space-time memory read block, and a decoder. The query encoder takes as input the current frame without a mask, and also produces key and value embeddings. The space-time memory read block then uses these embeddings for the current frame to produce feature maps by attending to the memory. The resulting feature maps are processed by the decoder to produce the segmentation masks for the current frame.

STM-VOS is first pre-trained on simulated data created from static image datasets, and afterwards the main training is done on a densely labeled video object segmentation dataset like YouTube-VOS.
As baseline, we perform the main training of STM-VOS without any change in the training procedure, except for replacing the hand-labeled ground truth with the generated pseudo-labels. 
Although Box2Seg generally performs well, there will still be some errors in the pseudo-labels. For an effective training, it is important to limit the adverse effect of these errors. To this end, we evaluate two extensions for improving the results.
\PAR{Filtering out Bad Masks.}
One possibility to deal with errors in the labels is to manually filter out bad masks. This does require manual effort, but is significantly cheaper than labeling all masks by hand. 
We evaluate how suitable this technique is by simulating this process on YouTube-VOS. To this end, we calculate the intersection-over-union (IoU) of the pseudo-label with the ground truth mask and mark its quality as bad if the IoU is below a certain threshold. 
In order to disregard the bad masks during training, in the pseudo-labels we remove this mask and instead fill the bounding box of the object with an ignore label, for which during training no loss is applied.

\PAR{Robust Loss Function.}
An alternative to limit the influence of errors in the training annotations, is to use a more robust loss function. %
In the default setup, STM-VOS is trained by applying a pixel-wise cross entropy loss, \ie
\begin{equation}
l_{\theta}(x,y)=-\log p_{\theta}(x,y),
\end{equation}
where $x$ is the input, $y$ is the label for the considered pixel, and $\theta$ are the parameters to be optimized. For this loss function, the influence of a single pixel is unbounded and loss values can increase dramatically for wrongly-labeled data.
In the context of learning classification models with label noise, Menon \etal~\cite{Menon20ICLR} propose the partially Huberised cross entropy loss
\begin{equation}
\tilde{l}_{\theta}(x,y)=\begin{cases}
-\tau\cdot p_{\theta}(x,y)+\log\tau+1 & \mathrm{if}\ p_{\theta}(x,y)\leq\frac{1}{\tau}\\
-\log p_{\theta}(x,y) & \mathrm{else},
\end{cases}
\end{equation}
which asymptotically saturates so that the influence of wrongly labeled data is bounded and the result for training with noisy labels can be dramatically improved. The hyperparameter $\tau$ should be set according to the expected amount of noise in the labels. 
We adapt this loss to the task of VOS by applying it to every pixel individually with $\tau=3$ and afterwards taking the average over spatial positions as usual.

\begin{table}
\setlength{\tabcolsep}{2pt}
\small
\begin{centering}
\begin{tabular}{ccccccc}
\toprule 
Dataset & \multicolumn{2}{c}{Videos} & Avg & Objects & Ann. & Categories\tabularnewline
 & train & val & len.(s) & / video & fps & \tabularnewline
\midrule
DAVIS 2017 & 60 & 30 & 3.4 & 1.7 & 20 & -\tabularnewline
YouTube-VOS 2018 & 3,471 & 474 & 4.5 & 2.3 & 6 & 94\tabularnewline
TAO (bound. boxes) & 500 & 988 & 36.8 & 5.9 & 1 & 833\tabularnewline
TAO-VOS & 500 & 126 & 36.7 & 5.9 & 1 & 361\tabularnewline
\bottomrule
\end{tabular}
\par\end{centering}
\caption{\label{tab:datasets}Statistics of the most popular VOS datasets DAVIS and YouTube-VOS together with TAO (annotated only with bounding boxes) and the newly introduced TAO-VOS annotations. Sequences in TAO-VOS are significantly longer, cover more classes, and more objects per video than in previous VOS datasets. TAO-VOS has only 361 classes instead of 833, because we only annotated part of the validation set and we did not annotate the test set of TAO-VOS, and many classes only appear in the test set.}
\end{table}

\begin{figure*}
\begin{center}
\includegraphics[width=0.24\textwidth]{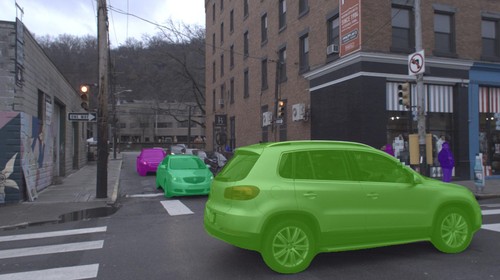}
\includegraphics[width=0.24\textwidth]{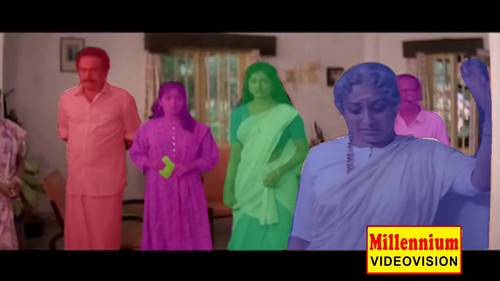}
\includegraphics[width=0.24\textwidth]{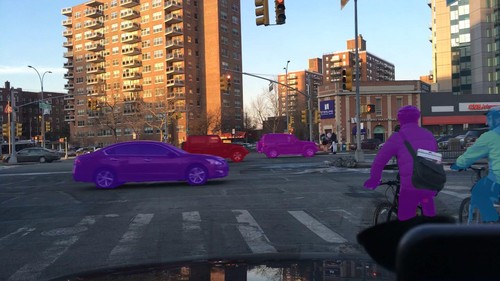}
\includegraphics[width=0.24\textwidth]{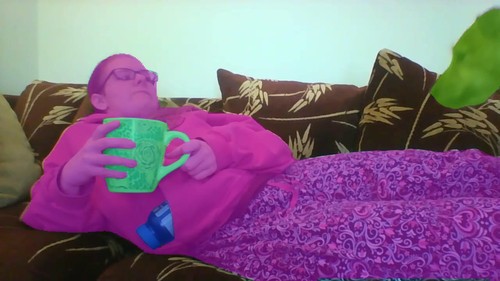}
\includegraphics[width=0.24\textwidth]{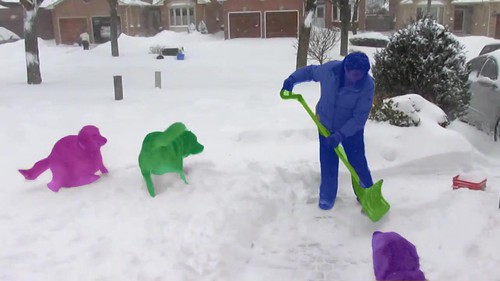}
\includegraphics[width=0.24\textwidth]{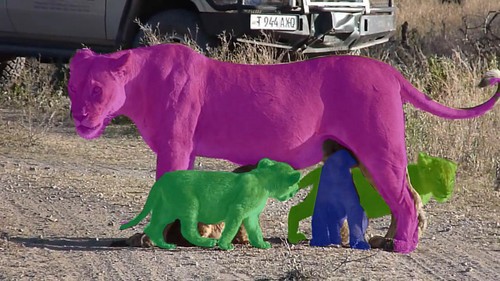}
\includegraphics[width=0.24\textwidth]{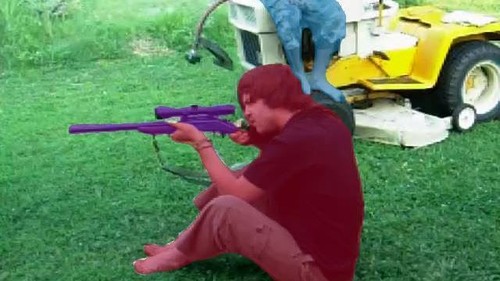}
\end{center}
\caption{\label{fig:taovos-dataset-vis-train}Example annotations of the TAO-VOS training set. We show one frame of one example sequence for each sub-dataset.}
\end{figure*}
\begin{figure*}
\begin{center}
\includegraphics[width=0.24\textwidth]{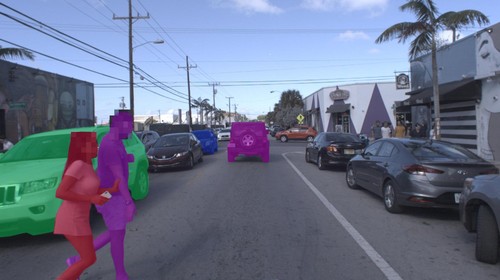}
\includegraphics[width=0.24\textwidth]{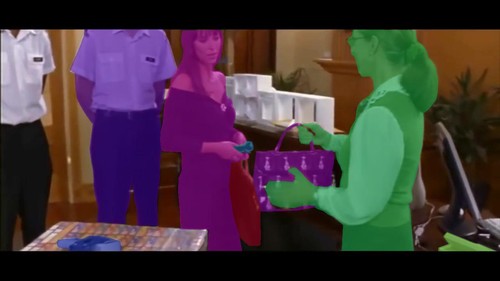}
\includegraphics[width=0.24\textwidth]{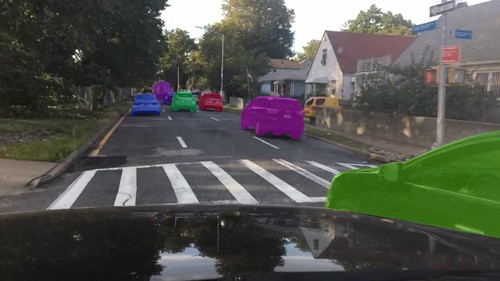}
\includegraphics[width=0.24\textwidth]{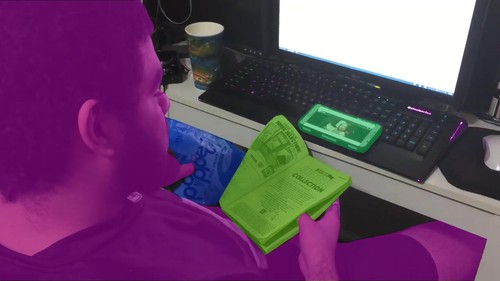}
\includegraphics[width=0.24\textwidth]{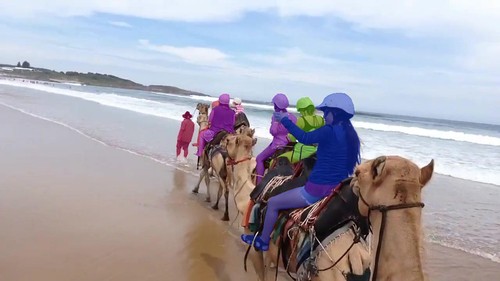}
\includegraphics[width=0.24\textwidth]{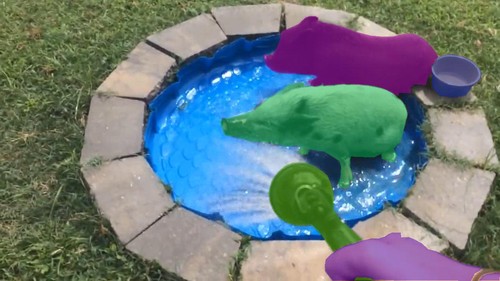}
\includegraphics[width=0.24\textwidth]{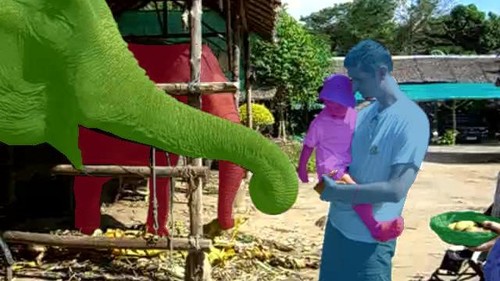}
\end{center}
\caption{\label{fig:taovos-dataset-vis-val}Example annotations of the TAO-VOS validation set. We show one frame of one example sequence for each sub-dataset.}
\end{figure*}

\begin{table*}
\begin{centering}
\begin{tabular}{ccccc}
\toprule 
Labels & $\mathcal{J}$ & recall@50\% & recall@70\% & Annotated masks (video)\tabularnewline
\midrule
Ground Truth & 100.0 & 100.0 & 100.0 & 159,976\tabularnewline
Box2Seg & 82.1 & 94.3 & 85.3 & 0\tabularnewline
Box2Seg + 1st frame FT. & 86.7 & 97.4 & 91.8 & 6,459\tabularnewline
Box2Seg + middle frame FT. & 87.6 & 97.9 & 93.1 & 6,459\tabularnewline
\bottomrule
\end{tabular}
\par\end{centering}
\caption{\label{tab:pseudo-label-quality}Quality of pseudo-labels together
with the number of used annotated ground truth masks on the YouTube-VOS training set. FT denotes fine-tuning, Annotated masks (video) refers to the
number of used annotated masks on video datasets (excluding static image
datasets), and recall@x\% denotes recall (in percent) at an IoU threshold of x\%.}
\end{table*}
\section{The TAO-VOS Benchmark}
Recently, the Tracking Any Object (TAO) \cite{Dave20Arxiv} benchmark for multi-object tracking has been introduced. It consists of 500 videos for training, 988 videos for validation, and 1,419 videos for testing. TAO features 833 different classes, and contains on average 5.9 objects per video. To ensure a high diversity of the data, TAO is composed of videos obtained from seven different existing video datasets: Charades \cite{Sigurdsson16ECCV}, ArgoVerse \cite{Chang19CVPR}, HACS \cite{Zhao19ICCV}, AVA \cite{Gu18CVPR}, YFCC100M \cite{Thomee16ACM}, BDD \cite{Yu20CVPR}, and LaSOT \cite{Fan19CVPRLASOT}.

Tab.~\ref{tab:datasets} shows statistics of the TAO dataset and compares it to the most important VOS datasets. It can be seen, that with 36.8 seconds, the average duration of the videos of TAO are much longer than in  DAVIS 2017 \cite{DAVIS2017} and YouTube-VOS \cite{Xu18Arxiv}. Additionally, the average number of objects per video is much higher. DAVIS does not use pre-defined categories and the number of videos is very small. With 833 categories, TAO covers much more diversity than YouTube-VOS which features 94 classes. Moreover, for DAVIS and YouTube-VOS, objects almost always occupy a large fraction of the video, while for TAO new objects often appear and disappear.

Due to these unique challenges, TAO has a strong potential to serve as a new VOS benchmark. However, due to its large size, it was only annotated with bounding boxes. Furthermore, TAO is annotated at a lower frame-rate than DAVIS and YouTube-VOS, but this is not a problem since most modern VOS methods sample a few frames with a random time gap for each training step. %
Moreover, not all objects in all videos are annotated in TAO, however this is also not a problem for VOS, where the objects of interest are specified by their first-frame ground truth masks.
In order to make this benchmark usable for VOS, we manually label the middle frame of each object of the training set with a mask, and then use Box2Seg with fine-tuning (\cf Sec.~\ref{sec:creating-pseudo-labels}) to generate masks for the whole training set of 500 sequences. Because TAO contains many more objects per sequence and is significantly more challenging than YouTube-VOS, the quality of the pseudo-labels is lower. Hence, in order to guarantee a high quality of the training set, we manually identified pseudo-labels with errors and added additional masks to fix the majority of these.
First, for the middle frames, we manually created 5,666 masks. Afterwards, we manually added another 9,680 masks to guarantee a high quality of pseudo-labels. In total, 12,513 of the total 59,200 masks (21\%) were annotated manually.

While the training set can be semi-automatically labeled, we argue that for evaluation, a definite ground truth without any mistakes is important. Hence, we select a representative subset (same amount from each sub-dataset) of 126 video sequences of the 988 validation videos and label each frame which has bounding box annotations by hand with masks to obtain the TAO-VOS validation set. In total, we annotated 14,987 masks for the validation set. %

Tab.~\ref{tab:datasets} shows statistics of the new TAO-VOS benchmark and Figures~\ref{fig:taovos-dataset-vis-train} and \ref{fig:taovos-dataset-vis-val} show examples of the annotations on the training and the validation set, respectively.

\section{Experiments}
\label{sec:experiments}
In the following, we perform experiments to validate the quality of the pseudo-labels, their effectiveness for training VOS models, and finally we present results of state-of-the-art VOS methods on the new TAO-VOS benchmark.

\subsection{Quality of Pseudo-Labels}

As our first experiment, we evaluate the quality of the pseudo-labels for the YouTube-VOS training set consisting of 3,471 sequences. Tab.~\ref{tab:pseudo-label-quality} shows the mean IoU $\mathcal{J}$ together with the number of used manually annotated masks for different sets of labels.
For a conventional VOS evaluation, also frames for which an object is not present are evaluated and the IoU is set to $1.0$ if also the predicted mask is empty. Since we assume that the ground truth boxes are given, here we take the average only over frames in which an object is present, which leads to lower scores but gives a more intuitive understanding of the label quality.

Without annotating any masks, a mean IoU of $82.1$ is reached. When fine-tuning using the middle frame, this result is improved by $5.5$ percentage points at the cost of annotating only 6,459 masks. It can also be seen that using the middle frame for fine-tuning is slightly more effective than using the first frame. Fig.~\ref{fig:pseudo-labels-qualitative} shows a qualitative example of the label quality without and with fine-tuning using the middle frame.

Another way to evaluate the labels is to define a quality requirement for each mask, \eg we require a mask to have at least 50\% or 70\% IoU with the ground truth mask and then we measure the fraction of automatically generated masks for which this quality is reached. The corresponding results are shown in Tab.~\ref{tab:pseudo-label-quality} as recall@50\% (recall at 50\% IoU threshold)  and as recall@70\%. It can be seen that without using any manual labels, already $94.3\%$ of the masks reach an IoU of at least 50\% and $85.3\%$ even reach an IoU of at least $70\%$. When fine-tuning using the middle frame, these values are largely improved to $97.9\%$ for the $50\%$ threshold and to $93.1\%$ for the 70\% threshold. %

\subsection{Training using Pseudo-Labels}

\begin{table*}
\begin{centering}
\begin{tabular}{cccc}
\toprule 
Setup & Annotated Masks & DAVIS'17 val & YouTube-VOS'18 val\tabularnewline
 & for Videos & $\mathcal{J}\&\mathcal{F}$ & $\mathcal{J}\&\mathcal{F}$\tabularnewline
\midrule 
Fully-Supervised (YouTube-VOS) \cite{Oh19ICCV} & 159,976 & 78.6 & 79.4\tabularnewline
Fully-Supervised (DAVIS'17) \cite{Oh19ICCV} & 9,627 & 71.6 & 56.3\tabularnewline
Pre-training only \cite{Oh19ICCV} & 0 & 60.0 & 69.1\tabularnewline
\midrule
Fully-Supervised (ours) & 159,976 & 78.8 & 79.3\tabularnewline
Fully-Supervised (ours) + Robust Loss & 159,976 & 78.2 (-0.6) & 79.8 (+0.5)\tabularnewline
Box2Seg Labels & 0 & 74.9 (-3.9) & 74.3 (-5.0)\tabularnewline
Box2Seg labels + Filtering & {*} & 76.0 (-2.8) & 74.6 (-4.7)\tabularnewline
Box2Seg labels + Robust Loss & 0 & 76.1 (-2.7) & 76.2 (-3.1)\tabularnewline
Box2Seg Fine-tuned labels & 6,459 & \textbf{77.7 (-1.1)} & 77.3 (-2.0)\tabularnewline
Box2Seg fine-tuned labels + Robust Loss & 6,459 & 76.9 (-1.9) & \textbf{77.9 (-1.4)}\tabularnewline
\bottomrule
\end{tabular}
\par\end{centering}
\caption{\label{tab:youtube-vos-results}Results of STM-VOS when training with
different labels. Fully-Supervised (DAVIS'17): training on DAVIS 2017 instead of YouTube-VOS. Pre-training only: training only on static image datasets. Fully-Supervised (ours): our re-produced result using the code from STM-VOS \cite{Oh19ICCV}. Box2Seg labels + Filtering: using the pseudo-labels from Box2Seg and replacing low-quality masks with an ignore label (\cf Sec.~\ref{sec:train-with-pseudo-labels}). {*}: indicates manual effort to assess
the quality of the masks. Robust Loss: use of partially Huberised cross entropy loss (\cf Sec.~\ref{sec:train-with-pseudo-labels}). Box2Seg fine-tuned labels: using the labels from Box2Seg which has been fine-tuned on the middle frame (\cf Sec.~\ref{sec:creating-pseudo-labels}).}
\end{table*}

Now we want to investigate how the label quality translates to the final VOS performance obtained when using these labels for training. Tab.~\ref{tab:youtube-vos-results} shows the results together with the number of used annotated masks on video datasets. For each configuration, first STM-VOS is pre-trained using static image datasets, and afterwards further trained on YouTube-VOS using the specified set of labels.

The results of the first three lines are directly copied from the STM-VOS paper \cite{Oh19ICCV}, and indicate that training only using static images leads to poor results. Additionally, when adding the small but densely-annotated DAVIS 2017 training set, the results on the DAVIS 2017 validation set improve, but the results on the YouTube-VOS 2018 validation set degrade due to over-fitting.

For our own experiments, we use the code and the initialization checkpoint from static image pre-training provided by the authors. We then re-implemented the missing training code, and re-ran the training with full supervision and closely matched the results provided by the authors. We further show that the robust loss function does not lead to a significant change when applied with full supervision.

When training using the Box2Seg labels which require 0 annotated masks on video datasets, we achieve a $\mathcal{J}\&\mathcal{F}$ score of $74.9$ on the DAVIS 2017 validation set and $74.3$ on the YouTube-VOS 2018 validation set, respectively. This is only 3.9 percentage points less than our fully-supervised version on DAVIS and 5.0 percentage points less on YouTube-VOS, albeit the effort to annotate 159,976 segmentation masks was saved. 
Note that using the Box2Seg labels performs significantly better than both only doing static-image pre-training and to train using the fully-annotated DAVIS 2017 train set.

When manually filtering out bad masks (\cf Sec.~\ref{sec:train-with-pseudo-labels}), this gap reduced further by 1.1 percentage points on DAVIS, and by 0.3 percentage points on YouTube-VOS. However, manual effort needs to be spent to assess the quality of the masks, which seems not worth it for the small improvement.
We find that using the robust loss function (\cf Sec.~\ref{sec:train-with-pseudo-labels}) instead is more effective (1.2 and 1.9 percentage points improvement for DAVIS and YouTube-VOS, respectively), and does not require any manual effort.

When annotating the 6,459 middle frame masks (roughly $4\%$ of all masks) for fine-tuning Box2Seg, %
the gap to full supervision is greatly reduced to a small difference of 1.1 percentage points and 2.0 percentage points for DAVIS 2017 and YouTube-VOS, respectively. 
Using the labels obtained with fine-tuning, because of the higher label quality, the effect of the robust loss function is less significant. The robust loss slightly improves the results on YouTube-VOS, while slightly degrading the results on DAVIS.

In the supplemental material, we show that training using Box2Seg labels is also effective for the related task of video instance segmentation \cite{Yang19ICCV}.

\begin{table*}
\small
\setlength{\tabcolsep}{2.5pt}
\centering{}%
\begin{tabular}{cccccc}
\toprule 
Setup & Ann. Masks & Ann. Boxes & Ann. Masks & DAVIS'17 val & YouTube-VOS'18 val\tabularnewline
 & for Images & for Videos & for Videos & $\mathcal{J}\&\mathcal{F}$ & $\mathcal{J}\&\mathcal{F}$\tabularnewline
\midrule
STM-VOS (Box2Seg labels + Robust Loss) & \ding{51} & \ding{51} & 0 & 76.1 & 76.2\tabularnewline
STM-STM (Box2Seg Fine-tuned Labels) & \ding{51} & \ding{51} & 6,459 & 77.7 & 77.3\tabularnewline
\midrule
PReMVOS \cite{Luiten18ACCV} & \ding{51} & \ding{51} & 0 & 77.8 & 66.9\tabularnewline
Siam R-CNN \cite{Voigtlaender20CVPR} & \ding{51} & \ding{51} & 0 & 70.6 & 68.3\tabularnewline
\midrule
MAST \cite{Lai20CVPR} & \ding{55} & \ding{55} & 0 & 65.5 & 64.2\tabularnewline
MuG \cite{Lu20CVPR} & \ding{55} & \ding{55} & 0 & 56.1 & -\tabularnewline
UVC \cite{Li19NIPS} & \ding{55} & \ding{55} & 0 & 59.5 & -\tabularnewline
CorrFlow \cite{Lai19BMVC} & \ding{55} & \ding{55} & 0 & 50.3 & 46.6\tabularnewline
\bottomrule
\end{tabular}\caption{\label{tab:comparison-weakly}Comparison of our results to other methods
for weakly-supervised VOS. Here, Ann. Masks and Ann. Boxes denote the use of annotated masks and bounding boxes, respectively.
Especially on YouTube-VOS, there is a big gap %
to the results of all previous methods.}
\end{table*}

\subsection{Comparison to Weakly-Supervised Methods}
In addition to interpreting our approach as a method for reducing the annotation effort, the combination of pseudo-label generation by Box2Seg and then training STM-VOS on these labels can also be seen as a weakly-supervised method for VOS. In Tab.~\ref{tab:comparison-weakly} we compare the results of this approach to other weakly-supervised VOS methods. We are only aware of two other methods, PReMVOS  \cite{Luiten18ACCV}, and Siam R-CNN \cite{Voigtlaender20CVPR}, which consider the training setup of only using bounding box annotations for videos, but making use of segmentation masks for static image datasets.

On DAVIS 2017, PReMVOS slightly outperforms our approach, but this comes at the cost of having to use four different neural networks and a run-time of more than 37 seconds per frame (compared to 0.32 seconds per frame for STM-VOS). For YouTube-VOS, PReMVOS has a significant gap of almost 8 percentage points to our approach. STM-VOS with Box2Seg labels (0 segmentation annotations for videos required), achieves a $\mathcal{J}\&\mathcal{F}$ score of 76.2, while the closest result is by Siam R-CNN \cite{Voigtlaender20CVPR} with 68.3.

We also compare to a number of approaches which use even weaker supervision: MuG \cite{Lu20CVPR} uses ImageNet classification labels and unlabeled videos, while MAST \cite{Lai20CVPR}, UVC \cite{Li19NIPS}, and CorrFlow \cite{Lai19BMVC} are learned only from unlabeled videos. While these methods address an interesting research topic and achieve impressive results given that they do not need any annotations, the overall result quality is still far away from what can be achieved by including static-image training data with segmentation labels.

\subsection{TAO-VOS}
\label{sec:tao-vos-experiments}
\begin{table}[t]
\begin{centering}
\begin{tabular}{cccc}
\toprule 
Method & \multicolumn{1}{c}{TAO-VOS} & DAVIS'17 & YT-VOS'18\tabularnewline
 & $\mathcal{J}\&\mathcal{F}$ & $\mathcal{J}\&\mathcal{F}$ & $\mathcal{J}\&\mathcal{F}$\tabularnewline
\midrule
STM \cite{Oh19ICCV} & 63.8 & 81.8 & 79.4\tabularnewline
CFBI \cite{Yang20ECCV} & 66.3 & 81.9 & 81.4\tabularnewline
LWL \cite{Goutam20ECCV} & 59.5 & 81.6 & 81.5\tabularnewline
\bottomrule
\end{tabular}
\par\end{centering}
\centering{}\caption{\label{tab:tao-vos-results}Results on the validation sets of TAO-VOS,
DAVIS 2017, and YouTube-VOS. The result for STM-VOS for DAVIS here
includes fine-tuning on the DAVIS 2017 training set and all results on TAO-VOS include fine-tuning on the TAO-VOS training set.}
\end{table}

\begin{table}[t]
\begin{centering}
\begin{tabular}{cccc}
\toprule 
Method & \multicolumn{3}{c}{TAO-VOS val}\tabularnewline
 & $\mathcal{J}\&\mathcal{F}$ & $\mathcal{J}$ & $\mathcal{F}$\tabularnewline
\midrule
STM \cite{Oh19ICCV} (YouTube-VOS) & 62.3 & 59.9 & 64.7\tabularnewline
+ TAO-VOS fine-tuning & 63.8 (+1.5) & 61.5 & 66.1\tabularnewline
\midrule
CFBI \cite{Yang20ECCV} (YouTube-VOS) & 63.9 & 61.9 & 66.0\tabularnewline
+ TAO-VOS fine-tuning & 66.3 (+2.4) & 63.8 & 68.8\tabularnewline
\midrule
LWL \cite{Goutam20ECCV} (YouTube-VOS) & 55.9 & 53.0 & 58.7\tabularnewline
+ TAO-VOS fine-tuning & 59.5 (+3.6) & 56.7 & 62.4\tabularnewline
\bottomrule
\end{tabular}
\par\end{centering}
\centering{}\caption{\label{tab:taovos-finetuning}Effect of fine-tuning on the TAO-VOS training set. %
}
\end{table}

\begin{figure*}
\newcommand{\mywidth}{0.18}
STM\hspace{0.2mm}
\raisebox{-.42\height}{
\includegraphics[width=\mywidth\textwidth]{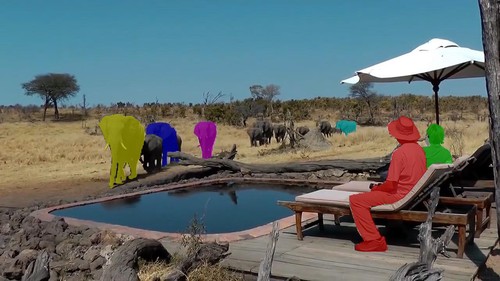}
\includegraphics[width=\mywidth\textwidth]{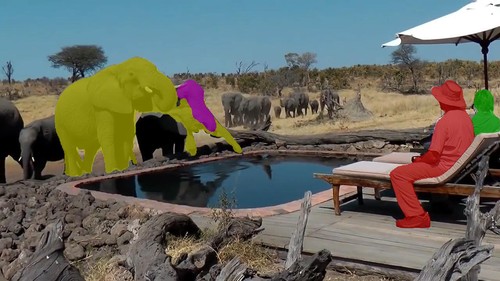}
\includegraphics[width=\mywidth\textwidth]{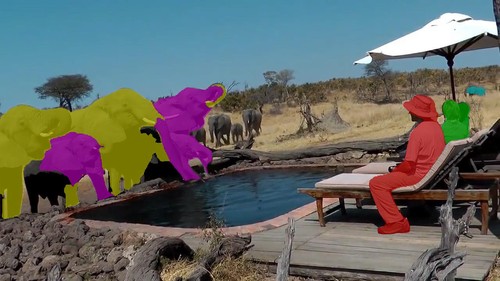}
\includegraphics[width=\mywidth\textwidth]{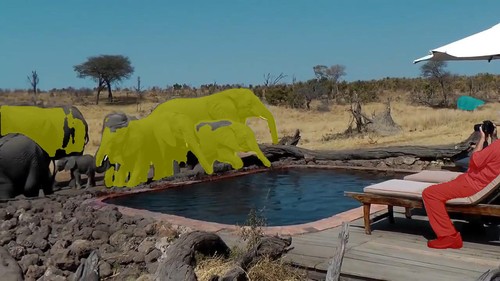}
\includegraphics[width=\mywidth\textwidth]{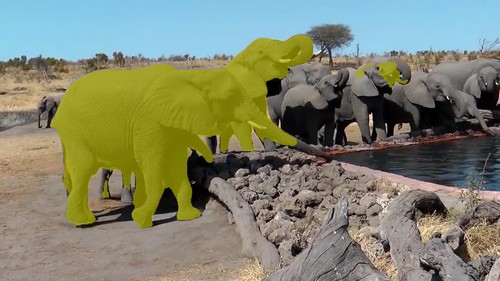}
}

CFBI\hspace{-0.5mm}
\raisebox{-.42\height}{
\includegraphics[width=\mywidth\textwidth]{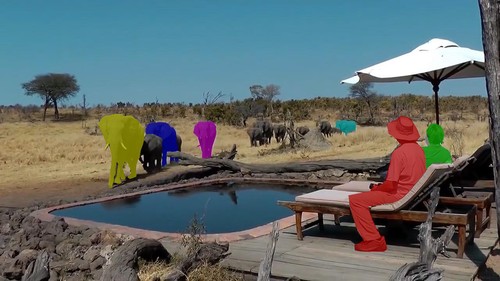}
\includegraphics[width=\mywidth\textwidth]{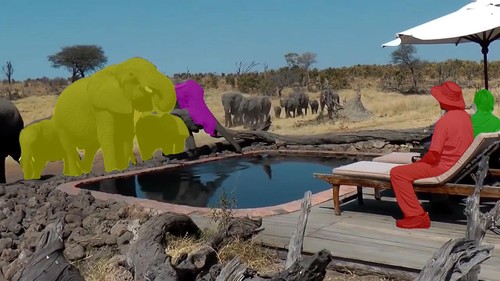}
\includegraphics[width=\mywidth\textwidth]{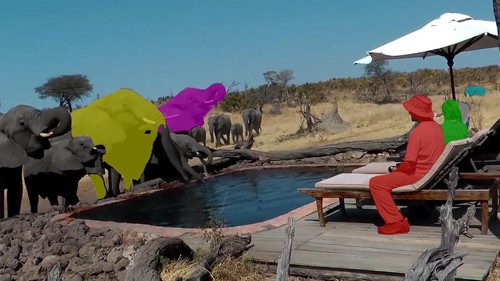}
\includegraphics[width=\mywidth\textwidth]{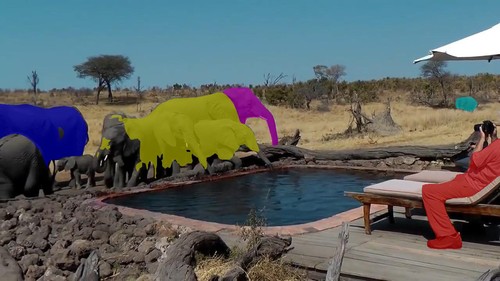}
\includegraphics[width=\mywidth\textwidth]{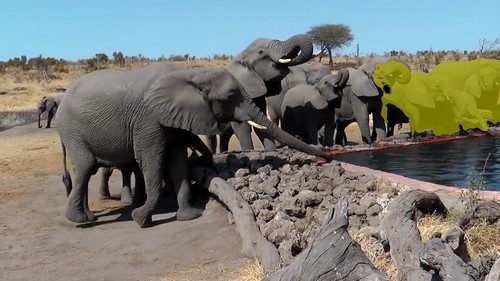}
}

LWL
\raisebox{-.42\height}{
\includegraphics[width=\mywidth\textwidth]{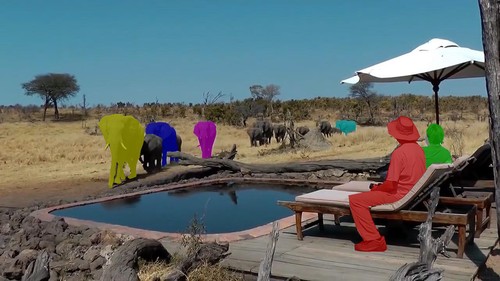}
\includegraphics[width=\mywidth\textwidth]{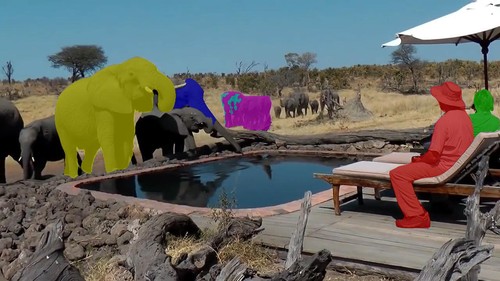}
\includegraphics[width=\mywidth\textwidth]{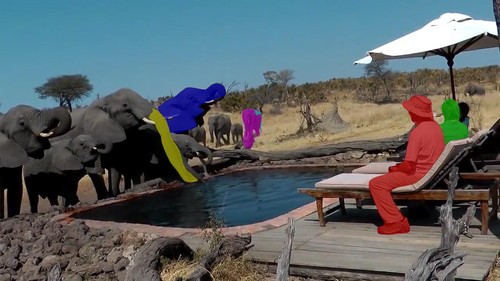}
\includegraphics[width=\mywidth\textwidth]{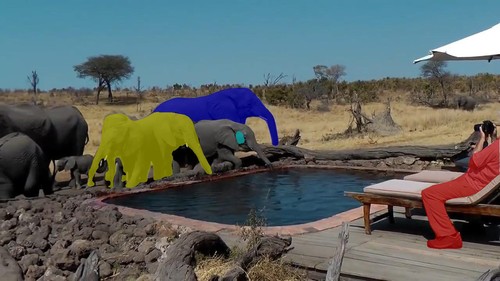}
\includegraphics[width=\mywidth\textwidth]{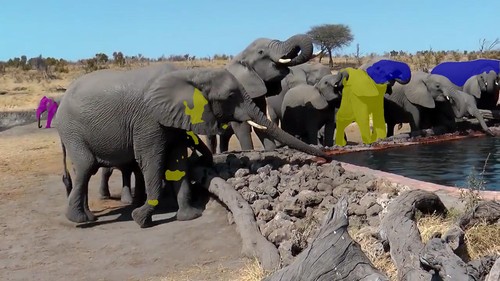}
}

GT\hspace{2.7mm}
\raisebox{-.42\height}{
\includegraphics[width=\mywidth\textwidth]{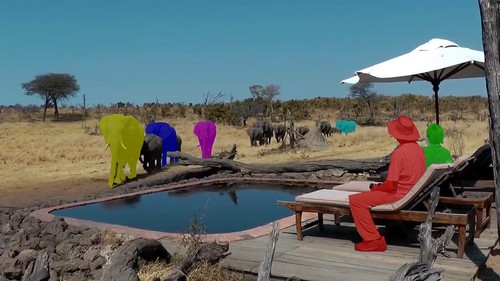}
\includegraphics[width=\mywidth\textwidth]{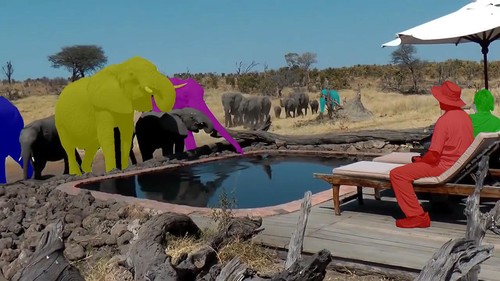}
\includegraphics[width=\mywidth\textwidth]{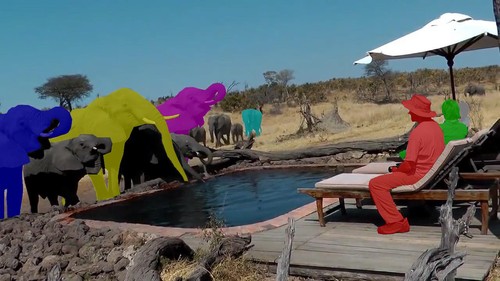}
\includegraphics[width=\mywidth\textwidth]{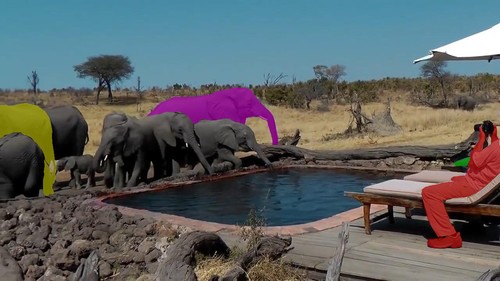}
\includegraphics[width=\mywidth\textwidth]{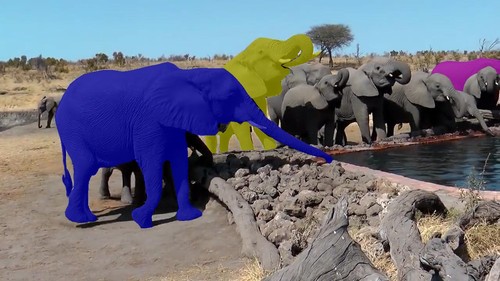}
}

STM\hspace{0.2mm}
\raisebox{-.42\height}{
\includegraphics[width=\mywidth\textwidth]{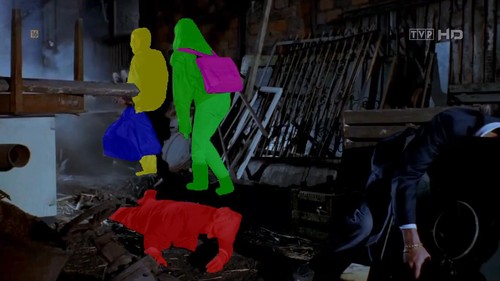}
\includegraphics[width=\mywidth\textwidth]{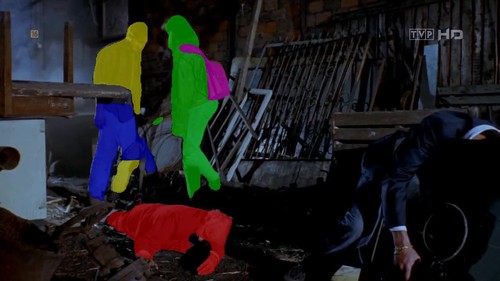}
\includegraphics[width=\mywidth\textwidth]{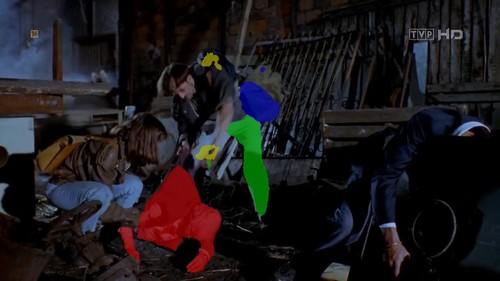}
\includegraphics[width=\mywidth\textwidth]{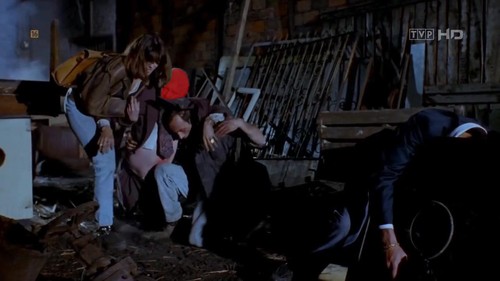}
\includegraphics[width=\mywidth\textwidth]{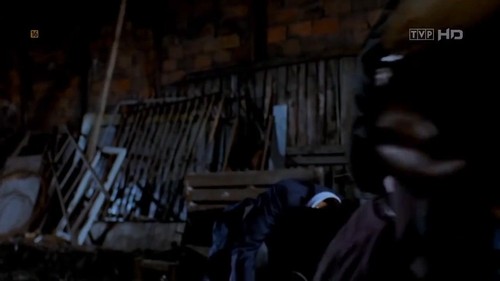}
}

CFBI\hspace{-0.5mm}
\raisebox{-.42\height}{
\includegraphics[width=\mywidth\textwidth]{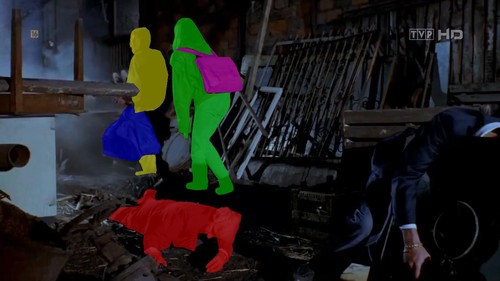}
\includegraphics[width=\mywidth\textwidth]{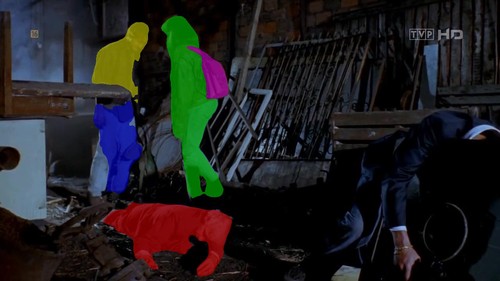}
\includegraphics[width=\mywidth\textwidth]{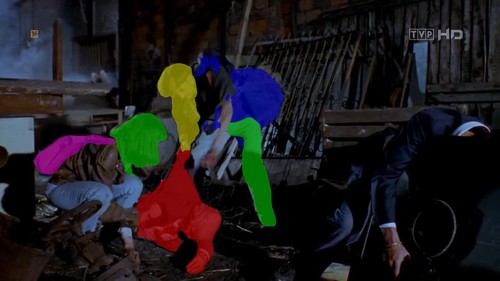}
\includegraphics[width=\mywidth\textwidth]{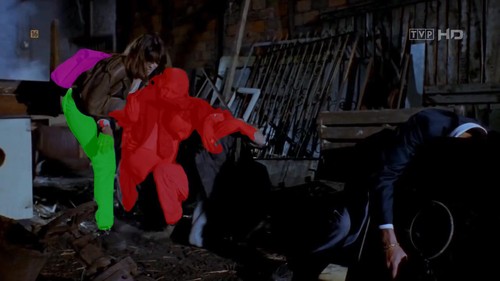}
\includegraphics[width=\mywidth\textwidth]{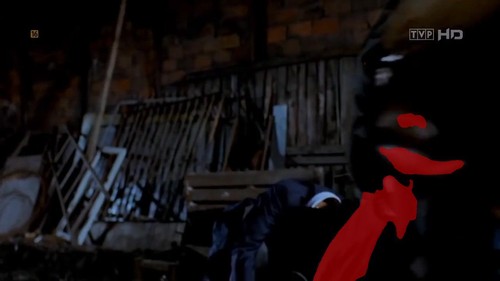}
}

LWL
\raisebox{-.42\height}{
\includegraphics[width=\mywidth\textwidth]{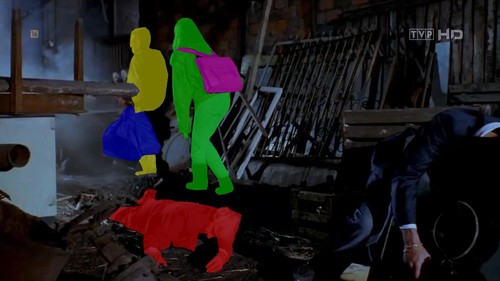}
\includegraphics[width=\mywidth\textwidth]{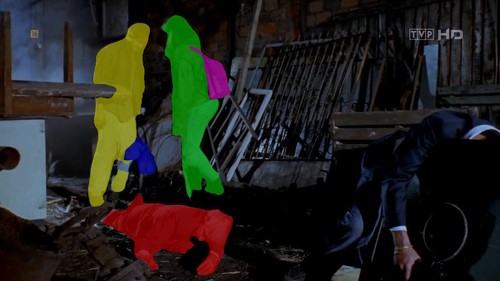}
\includegraphics[width=\mywidth\textwidth]{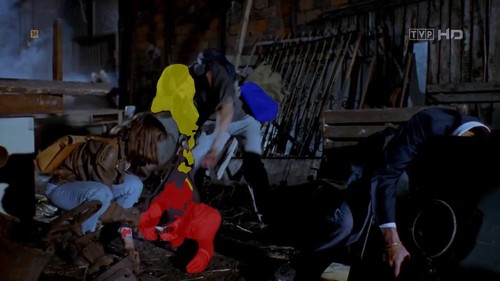}
\includegraphics[width=\mywidth\textwidth]{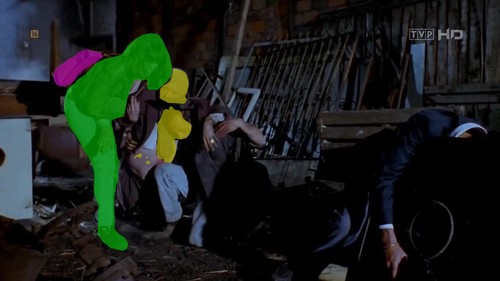}
\includegraphics[width=\mywidth\textwidth]{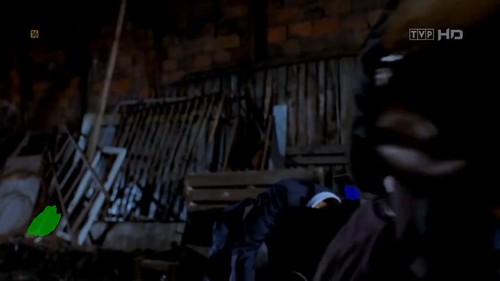}
}

GT\hspace{2.7mm}
\raisebox{-.42\height}{
\includegraphics[width=\mywidth\textwidth]{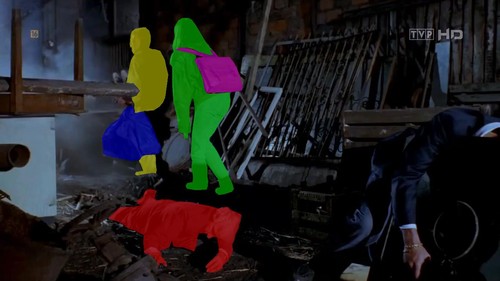}
\includegraphics[width=\mywidth\textwidth]{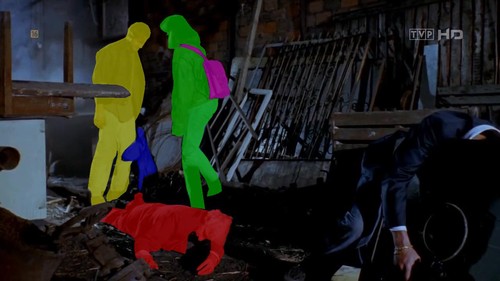}
\includegraphics[width=\mywidth\textwidth]{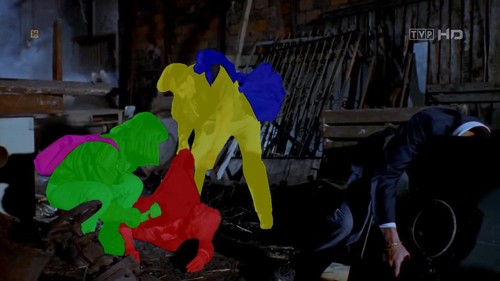}
\includegraphics[width=\mywidth\textwidth]{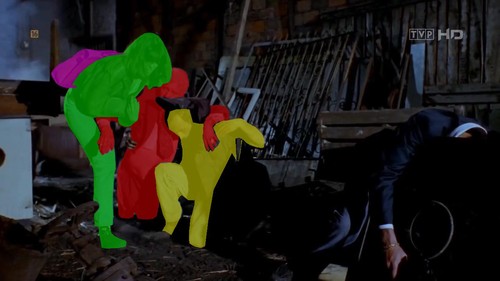}
\includegraphics[width=\mywidth\textwidth]{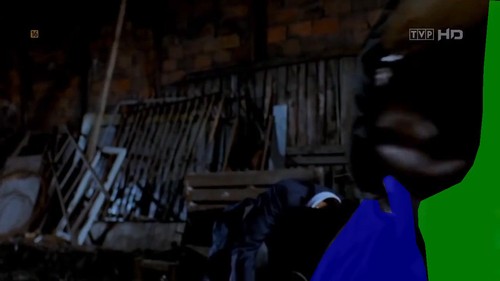}
}

\caption{\label{fig:taovos-qualitative}Qualitative results on the TAO-VOS validation set. Here we compare STM-VOS \cite{Oh19ICCV}, CFBI \cite{Yang20ECCV}, LWL \cite{Goutam20ECCV}, and the ground truth (GT). For the two considered hard sequences from TAO-VOS, all considered VOS methods fail to segment the objects accurately after some time.}
\end{figure*}

Next we evaluate recent VOS methods on the new TAO-VOS benchmark. %
We consider three methods that together cover the current state-of-the-art: STM-VOS \cite{Oh19ICCV}, CFBI \cite{Yang20ECCV}, and LWL \cite{Goutam20ECCV}. CFBI employs a global and a local matching strategy to match the pixels of the current frame to pixels of other frames in order to transfer information which is used as internal guidance of the network to produce masks. Additionally, a special handling of the background and an instance-level attention mechanism are used. LWL \cite{Yang20ECCV} uses meta-learning to learn a target model of the object of interest. The target model predicts an encoding of the target mask, which is used by a decoder to produce segmentation masks. All three approaches are state-of-the-art VOS  methods and achieve excellent and very similar performance numbers on DAVIS and YouTube-VOS.

The TAO-VOS training set only consists of 500 sequences which is still much smaller than YouTube-VOS with 3,471 sequences. Hence, we initialize from YouTube-VOS pre-training and use TAO-VOS for fine-tuning (see supplemental material for implementation details).

The TAO-VOS validation set is only annotated at 1 frame per second (FPS), while the video data is available at around 30 FPS. Similar to the evaluation protocol for YouTube-VOS, an algorithm can either run only on the annotated frames, or it can use all frames. Intuitively, tracking objects at a higher frame-rate should be easier. %
For each method we tried both variants and report the best of the two results. Interestingly, we found that only LWL can benefit from the higher frame rate, while STM and CFBI produces better results when run only on the annotated frames (see supplemental material for details).

Tab.~\ref{tab:tao-vos-results} shows the results on TAO-VOS after fine-tuning, and we also show the results of the original methods on the validation sets of DAVIS 2017 and YouTube-VOS 2018 as a reference. It can be noted that the absolute performance on TAO-VOS is much lower than on DAVIS and YouTube-VOS. %
Furthermore, all three methods perform very similar on DAVIS and YouTube-VOS, while on TAO-VOS the result for LWL is significantly worse than the results of the two other methods.
These results show that TAO-VOS is much more challenging than DAVIS and YouTube-VOS, and that it can further reveal differences between methods which were not visible on existing datasets.
The best result of 66.3 $\mathcal{J}\&\mathcal{F}$ is relatively low and leaves much room for improvement by further methods, while the performance on DAVIS and YouTube-VOS shows strong signs of saturating.

In Tab.~\ref{tab:taovos-finetuning}, we analyze the effect of fine-tuning on the TAO-VOS training set. It can be seen that all three methods benefit from fine-tuning with the largest improvement of 3.6 percentage points achieved by LWL.%

Fig.~\ref{fig:taovos-qualitative} shows qualitative results on two example sequences. %
It can be seen that both considered sequences are very hard with many objects and that all three considered methods lose track of all objects after some time. Again, this highlights that TAO-VOS as a benchmark has strong potential to enable further progress in VOS.

\vspace{-0.1mm}
\section{Conclusion}
We started with the observation that the progress of VOS is strongly driven by the availability of VOS datasets, and that current VOS datasets are still relatively small and do not cover diverse, long, and challenging enough videos. Subsequently, we identified the high manual effort of dense pixel-wise annotations as a major bottleneck for further progress in VOS. Towards a solution, we show that dense labels might not be necessary, and only bounding boxes plus few or no segmentation annotations are enough to create pseudo-labels from which VOS methods can learn almost as much as from costly dense annotations. As a proof of concept, we labeled the TAO-VOS training set semi-automatically with high quality. %
In contrast to VOS performance saturating on current VOS benchmarks, the challenging TAO-VOS validation set reveals differences between and shortcomings of state-of-the-art VOS methods. 
We hope that our work inspires the creation of even larger VOS datasets which can be labeled with less effort.

\footnotesize \PARbegin{Acknowledgements:} 
This project has been funded, in parts, by 
ERC Consolidator Grant DeeViSe (ERC-2017-COG-773161), a Google Faculty Research Award, NSFC project Grant No.~U1833101, Shenzhen Science and Technologies project under Grant No.~JCYJ20190809172201639 and the Joint Research Center of Tencent and Tsinghua. 
The authors would like to thank Berin Gnana and Sabarinath Mahadevan for helpful discussions.
\normalsize

{\small
\bibliographystyle{ieee_fullname}
\bibliography{abbrev_short,paper}
}

\newpage
\vskip .375in
\twocolumn[\begin{center}
{\Large \bf Supplemental Material for\\Reducing the Annotation Effort for Video Object Segmentation Datasets \par}
\vskip 1.5em
\end{center}]
\appendix

\normalsize

\section{Video Instance Segmentation Experiments}
Video Instance Segmentation (VIS) \cite{Yang19ICCV} is a recently introduced computer vision task related to VOS. Unlike VOS, for VIS a pre-defined set of categories is used and no first-frame ground truth mask is given. All objects of the pre-defined categories need to be detected, segmented, classified into the right category, and tracked over time. The performance is measured by the Average Precision ($\mathcal{AP}$), and Average Recall ($\mathcal{AR}$) measures which are also used for image instance segmentation. However, for VIS, these measures are calculated on the whole track level (for details see \cite{Yang19ICCV}).

For VIS, so far the only dataset is YouTube-VIS \cite{Yang19ICCV} which features 40 categories, 131k instance masks, and 2,883 videos. In order to show that our method for creating pseudo-labels is not only effective for the VOS task, we also applied it to VIS, and used these labels for training STEm-Seg \cite{AtharMahadevan20ECCV}, a state-of-the-art VIS method.
STEm-Seg uses a single-stage network to learn a spatio-temporal pixel embedding to cluster instances over an entire video clip. We used the code provided by the authors to train STEm-Seg on our Box2Seg labels created for YouTube-VIS. The results are shown in Table~\ref{tab:youtubevis}. It can be seen that when using the labels created by Box2Seg based on only the bounding boxes of YouTube-VIS, the $\mathcal{AP}$ decreases from 34.6 to 30.5, \ie by 4.1 percentage points. When adding only a single manually annotated mask per object, and using it for fine-tuning Box2Seg as done in the main paper for YouTube-VOS, the gap decreases significantly to only 1.3 percentage points. This demonstrates that our pseudo-label generation method is also applicable for different datasets, tasks, and methods trained on them.

\begin{table}
\begin{centering}
\begin{tabular}{cccc}
\toprule 
Method & \multicolumn{3}{c}{TAO-VOS val}\tabularnewline
 & $\mathcal{J}\&\mathcal{F}$ & $\mathcal{J}$ & $\mathcal{F}$\tabularnewline
\midrule
STM \cite{Oh19ICCV}, annotated frames & 63.8 & 61.5 & 66.1\tabularnewline
STM \cite{Oh19ICCV}, all frames & 60.6 (-3.2) & 58.9 & 62.3\tabularnewline
\midrule
CFBI \cite{Yang20ECCV}, annotated frames & 66.3 & 63.8 & 68.8\tabularnewline
CFBI \cite{Yang20ECCV}, all frames & 58.7 (-7.6) & 56.6 & 60.8\tabularnewline
\midrule
LWL \cite{Goutam20ECCV}, annotated frames & 55.6 & 53.5 & 57.8\tabularnewline
LWL \cite{Goutam20ECCV}, all frames & 59.5 (+3.9) & 56.7 & 62.4\tabularnewline
\bottomrule
\end{tabular}
\par\end{centering}
\centering{}\caption{\label{tab:taovos-allframes}Effect of doing inference only at the
annotated frames (1 frame per second), or evaluating on all frames.
Only LWL benefits from the higher frame rate, and STM and CFBI perform
better with the lower frame rate.}
\end{table}

\section{Effect of Inference Frame Rate}
The TAO-VOS validation set is only annotated at 1 frame per second, while the video data is available at a higher frame rate of around 30 frames per second. Similar to the evaluation protocol for YouTube-VOS, an algorithm can either run only on the annotated frames, or it can use all frames. Intuitively, tracking objects at a higher frame-rate should be easier, because there is less motion between two adjacent frames. For each method we tried both variants (see Table~\ref{tab:taovos-allframes}). Interestingly, only LWL can benefit from the higher frame rate, while STM-VOS and CFBI produced better results when run only on the annotated frames. The difference is especially strong for CFBI, where using all frames is 7.6 percentage points worse than when just evaluating on the annotated frames.
By qualitative inspection, we found that for STM-VOS and CFBI at a high frame rate the predicted mask can get slightly smaller from frame to frame or it can grow. In both cases, small errors accumulate over time and lead to worse results. When using the low frame rate, it is harder to transfer information between the frames, because there is more motion, but at the same time, small errors cannot easily accumulate over time. We found that LWL produces masks which are more stable over time, and hence it can benefit from the higher frame rate.

\section{Implementation Details of Fine-Tuning}
Here we provide implementation details of fine-tuning the three considered methods on the TAO-VOS training set. Note that we mainly mention the differences to the default settings of the provided code for the considered methods.

For fine-tuning STM-VOS we used the robust loss function, \ie the partially Huberised cross entropy loss described in the main paper. %
Fine-tuning was then done using the Adam optimizer \cite{adam} with a learning rate of $10^{-7}$ for 50 epochs (the main training on YouTube-VOS was done with Adam with a learning rate of $10^{-6}$ for 300 epochs).

CFBI and LWL use different loss functions which are not as straightforward to adapt for robustness, hence we keep their original loss functions. %

For CFBI, we fine-tuned on the TAO-VOS training set for 5,000 steps using a learning rate of $10^{-3}$ (the main training on YouTube-VOS used a learning rate of $10^{-2}$).

For LWL, in the default setup, the learning rate differs for different weights and training is done for 70 epochs. For fine-tuning on the TAO-VOS training set, we reduced the learning rate by a factor of 10 and fine-tuned for 20 epochs.

\begin{table*}
\begin{centering}
\begin{tabular}{cccccc}
\toprule 
\multirow{2}{*}{Method} & \multicolumn{5}{c}{YouTube-VIS validation set}\tabularnewline
 & $\mathcal{AP}$ & $\mathcal{AP}@50$ & $\mathcal{AP}@75$ & $\mathcal{AR}@1$ & $\mathcal{AR}@10$\tabularnewline
\midrule
STEm-Seg \cite{AtharMahadevan20ECCV}  & 34.6 & 55.8 & 37.9 & 34.4 & 41.6\tabularnewline
STEm-Seg (Box2Seg labels) & 30.5 (-4.1) & 50.9 & 33.5 & 29.8 & 36.4\tabularnewline
STEm-Seg (Box2Seg fine-tuned labels) & 33.3 (-1.3) & 53.1 & 36.0 & 33.1 & 39.5\tabularnewline
\bottomrule
\end{tabular}
\par\end{centering}
\centering{}\caption{\label{tab:youtubevis}Results for Video Instance Segmentation (VIS) on the YouTube-VIS validation set. Here we use the state-of-the-art method STEm-Seg and train it with different sets of labels. $\mathcal{AP}$@50 and $\mathcal{AP}$@75 denote average precision at an IoU threshold of 50\% or 75\%, respectively, and $\mathcal{AP}$ is averaged over different thresholds. $\mathcal{AR}$ denotes the maximum recall achieved by a fixed number of segmented instances per video \cite{Yang19ICCV}.}
\end{table*}

\end{document}